\documentclass{article}

\usepackage[dblblindworkshop,final]{neurips_2026}
\workshoptitle{Foundation Models for Temporal Systems}
\usepackage{enumitem}

\usepackage[utf8]{inputenc} 
\usepackage[T1]{fontenc}    
\usepackage[hidelinks]{hyperref} 
\usepackage{url}            
\usepackage{graphicx}       
\usepackage{booktabs}       
\usepackage{array}          
\usepackage{tabularx}       
\usepackage{amsmath,amssymb} 
\usepackage{amsfonts}       
\usepackage{nicefrac}       
\usepackage{microtype}      
\usepackage{xcolor}         
\usepackage[normalem]{ulem} 
\usepackage{booktabs}
\usepackage{multirow}
\usepackage[table]{xcolor}
\usepackage{wrapfig}
\usepackage[most]{tcolorbox}
\usepackage{listings}
\lstdefinestyle{promptstyle}{
    basicstyle=\ttfamily\small,
    breaklines=true,
    breakatwhitespace=false,
    columns=fullflexible,
    keepspaces=true,
    showstringspaces=false,
    aboveskip=0pt,
    belowskip=0pt,
    frame=none,
    xleftmargin=0pt,
    xrightmargin=0pt
}

\newtcblisting{promptbox}{
    listing only,
    listing options={style=promptstyle},
    breakable,
    colback=gray!3,
    colframe=gray!55,
    boxrule=0.5pt,
    arc=2mm,
    left=7pt,
    right=7pt,
    top=7pt,
    bottom=7pt,
    before skip=6pt,
    after skip=10pt
}

\newcommand{\ctx}[1]{{\color{black!55}#1}\par\smallskip}
\newcommand{\dec}[1]{\makebox[0pt][r]{$\blacktriangleright$\,}#1\par\smallskip}
\newcommand{\meth}[3]{\textbf{#1}\par\texttt{\footnotesize #2}\par\smallskip{\footnotesize #3}}

\title{Put the World Back: Timestamps Matter in Time Series Reasoning}

\author{%
  \begin{tabular}{@{}c@{\hspace{3em}}c@{}}
    Tingrui Huang\thanks{Equal contribution.} &
    Yuanyuan Deng\footnotemark[1] \\
    \normalfont Eindhoven University of Technology &
    \normalfont Eindhoven University of Technology \\
    \normalfont\texttt{t.huang2@student.tue.nl} &
    \normalfont\texttt{y.deng2@tue.nl} \\[8pt]
    Qingren Yao &
    Joaquin Vanschoren \\
    \normalfont Eindhoven University of Technology &
    \normalfont Eindhoven University of Technology \\
    \normalfont\texttt{q.yao@tue.nl} &
    \normalfont\texttt{j.vanschoren@tue.nl} \\[8pt]
    Mykola Pechenizkiy &
    Songgaojun Deng \\
    \normalfont Eindhoven University of Technology &
    \normalfont Eindhoven University of Technology \\
    \normalfont\texttt{m.pechenizkiy@tue.nl} &
    \normalfont\texttt{s.deng@tue.nl}
  \end{tabular}
}

\begin{document}

\maketitle

\begin{abstract}

Large language models (LLMs) are increasingly applied to time series reasoning. However, current evaluations often give models events in chronological order and already matched to the time series. This makes it unclear whether models use timestamps to connect events with observations or simply take shortcuts. To expose this ambiguity, we reformulate time series reasoning from a world-to-series projection perspective, viewing timestamps as explicit indices that connect numerical observations and external events to shared world states. We identify three structures that timestamps can make explicit: temporal causal structure, time semantics, and event--series alignment. Building on this perspective, we develop a diagnostic protocol that distinguishes timestamp-grounded reasoning from correlational shortcuts, a distinction that accuracy-based benchmarks cannot make. Controlled interventions on 310 questions from two tasks show that high accuracy can persist when timestamps are replaced by relative offsets or removed, while breaking event--series alignment induces systematic failures. 
When context is pre-aligned, models barely need timestamps. When order is missing, they use dates to recall world knowledge. Without dates, they rely on correlational shortcuts and rarely notice incorrect timestamps.
These findings show that high accuracy on pre-aligned, ordered inputs does not establish that a model uses timestamps to synchronize observations with the world.
We release our code, model outputs, and trajectory labels at \url{https://github.com/tingrui-huang/timestamps-matter}.

\end{abstract}

\section{Introduction}
\label{sec:introduction}

Large language models are increasingly used for time series reasoning such as explaining observed patterns, anticipating future behavior, and supporting decisions~\citep{kong2025achieving, chang2025survey}. Current evaluations predominantly treat the time series as a self-contained numerical object, with event context supplied as manually aligned text, paired with the sequence by the dataset designer~\citep{cai2024timeseriesexam, kong2025time, chen2025mtbench}. Although the setting has advanced time series reasoning development~\citep{kong2026timesage, guan2026timeomni}, it detaches the series from the timeline on which real-world reasoning actually operates.

We view time series as incomplete records of the real world, with timestamps linking numerical observations and events to when they happened. A value alone is a bare number, while a timestamped value can be associated with a particular world state. However, current benchmarks often provide event context in an already aligned and chronologically ordered form. 
A model may answer correctly without using timestamps. It might simply follow the given event order or match events with nearby text.
Consequently, benchmark accuracy alone cannot distinguish genuine temporal grounding from such shortcuts, and it remains unclear \textit{whether LLMs reason by linking observed dynamics to the events behind them}.

\begin{figure*}[t]
    \centering
    \includegraphics[width=0.7\textwidth]{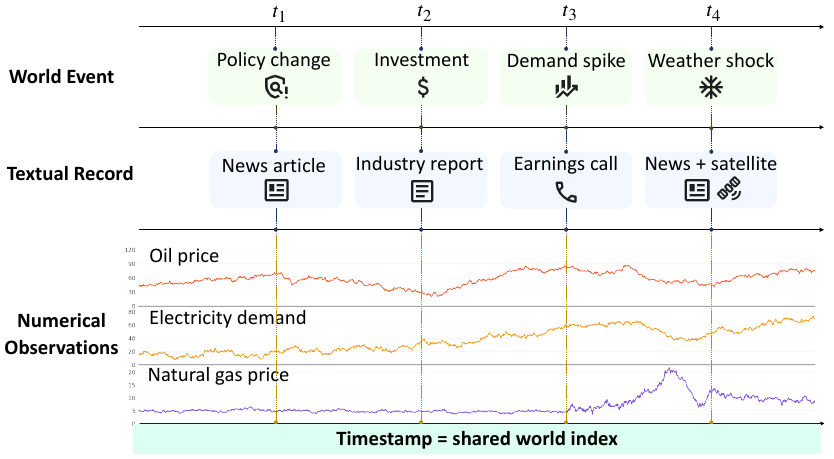}
    \vspace{-5pt}
    \caption{
    \textbf{Timestamps as a shared world index.}
    Real-world events are reflected in heterogeneous textual records and
    numerical observations. Timestamps provide a common index for aligning
    these records to the world states in which they occur.
    }
    \label{fig:world-index}
  \vspace{-10pt}
\end{figure*}

In this work, we recast time series reasoning as inference over a world-to-series projection (Figure~\ref{fig:world-index}), grounded in real-world temporal structure. We treat timestamps as the anchor bridging time series observations and the event sequence, exposing three structures including temporal causal structure, time semantics, and event--series alignment. 
Building on this view, we propose a diagnostic protocol that tests, for each structure, whether models actually use timestamps, rather than inferring temporal grounding from accuracy.
%
Our contributions are summarized as follows:
\begin{enumerate}[leftmargin=*]
    \item \textbf{A world-to-series projection perspective.} We reformulate time series reasoning as inference over a lossy world projection, with timestamps anchoring series values to world states.
    \item \textbf{A temporal causal diagnostic protocol.} We operationalize this perspective into a Temporal Causal Diagnostic Protocol (TCDP), pairing timestamp interventions with rules for what a trajectory must show to be grounded in time. TCDP reports which rules trajectories follow, which they violate, and which a benchmark never makes necessary because it supplies a shortcut such as event order.
    \item \textbf{Empirical findings and benchmark diagnosis.} 
    Applying TCDP to two TSRBench tasks, we show that accuracy on pre-aligned inputs hides whether timestamps are used. Models rely on the given event order instead of timestamps, and without it, they rarely check alignment.
\end{enumerate}

\section{Method}
\label{sec:method}

\subsection{Time Series as a World Projection}
\label{sec:framework}

We formally illustrate that a time series is not a self-contained numerical object but a lossy projection of a world process. Let $W$ denote a world process evolving over continuous time, approximated by a stream of textually recorded events $E = \{(\tau_j, e_j)\}_{j=1}^{m}$, and let $\pi$ be a projection that further reduces this process to one measurable quantity at discrete times. This can be expressed as
\begin{equation}
    W \approx E = \{(\tau_j, e_j)\}_{j=1}^{m}
    \xrightarrow{\;\pi\;}
    X = \{(t_i, x_i)\}_{i=1}^{n},
    \label{eq:projection}
\end{equation}
where the event records capture only part of the mechanisms, context, and intent behind $W$. The projection $\pi$ reduces this information further, leaving only numerical values. The timestamps $\tau_j$ and $t_i$ provide an explicit shared index back to $W$, a common timeline through which events and numerical observations can be associated with the world states in which they occur.
We identify three structures that a timestamp can make explicit when it is used as a world index. 

\textit{\textbf{(1) Temporal causal structure (why).}} Timestamps show how long an effect follows its cause and which events happen at the same time~\citep{granger1969investigating, runge2019detecting}. Both matter for causal claims, since a lag must fit a plausible mechanism and simultaneous events could each explain the effect. Order shows only which event came first. It cannot tell a one-hour lag from a one-week lag, and it places simultaneous events in sequence. A model that relies on sequence position thus reduces causal inference to \textit{post hoc ergo propter hoc}~\citep{eichler2013causal}.

\textit{\textbf{(2) Time semantics (when).}} Each $t_i$ is a key into world knowledge. The same pattern on different dates can reflect different world states and so needs different explanations~\citep{dhingra2022time}. 
Order shows only relative position, so it cannot provide this timestamp key.
A model without the key may read the pattern correctly yet attach it to the wrong world state.

\textit{\textbf{(3) Event--series alignment (what).}} Timestamps link events to observations. They show whether an event overlaps, precedes, or falls outside the observation window, and hence whether it is valid evidence~\citep{ning2018multi}. When both arrive pre-aligned, a model can link an event to the series simply because they appear close together in the text.

\subsection{Temporal Causal Diagnostic Protocol}
\label{sec:protocol}

\begin{figure}[t]
    \centering
    \includegraphics[width=\linewidth]{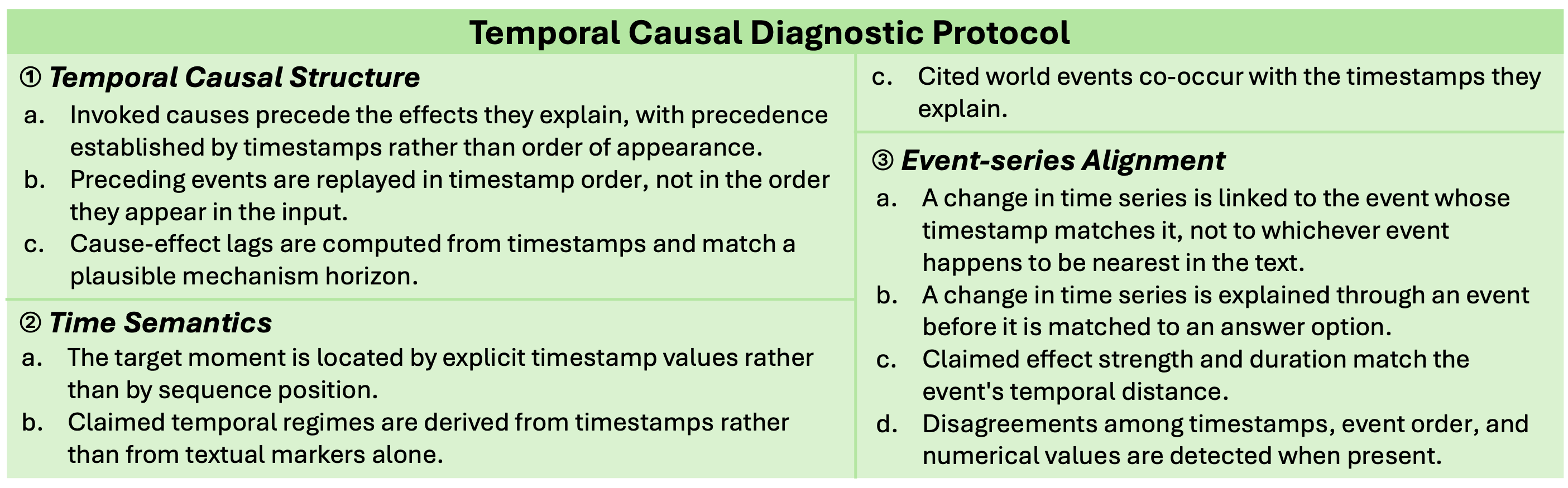}
    \vspace{-15pt}
    \caption{Checks in TCDP for auditing the temporal grounding of an LLM's reasoning trajectory.}
    \label{fig:tcdp}
  \vspace{-10pt}
\end{figure}

We operationalize the three structures as a Temporal Causal Diagnostic Protocol (TCDP), a checklist of binary rules applied to an LLM's reasoning \emph{trajectory}, grouped into three families (Figure~\ref{fig:tcdp}). Accuracy records \emph{whether} a conclusion is correct, while TCDP checks \emph{how} it is reached. 
The rules are necessary conditions for timestamp-grounded reasoning, not a score. 
For each rule, TCDP checks whether a benchmark requires it to answer correctly, and if so, whether trajectories follow it.


\section{Experiments}
\label{sec:experiments}

\textbf{Setup.}
We use two TSRBench tasks~\citep{yu2026tsrbench} (Appendix~\ref{app:tasks}).
In \emph{NBA} (150 questions), the model sees ordered play-by-play events with one event removed and a table of each team's score and win probability in which the missing row is marked. It chooses which of four candidate events fills the gap.
In \emph{GDELT} (160 questions), the model sees 6--8 news summaries, a GoldStein series (a $-10$ to $+10$ conflict--cooperation score per record), and the event times listed in random order and not paired with the summaries. It chooses the true event order from four options, so the task is designed to require the timestamps.

We apply three interventions to both tasks (Figure~\ref{fig:data_intervention}, Appendix~\ref{app:interventions}).
\textsc{Relative} gives each time as an offset from the target (NBA, seconds) or first record (GDELT, hours), keeping order and spacing but losing game or calendar time.
\textsc{No-TS} deletes all timestamps.
\textsc{Shuffled} leaves the series unchanged and gives each event a time that belongs to another observation. This breaks event--series alignment.
We evaluate \texttt{Sonnet-5}, \texttt{GLM-5.3-Flash}, \texttt{Qwen3.6-35B-A3B} and \texttt{Qwen3.5-9B} on NBA; \texttt{GLM-5.3-Flash} and \texttt{Sonnet-5} on GDELT. An LLM labels each GLM-5.3-Flash trajectory with the method deciding its answer (Appendix~\ref{app:labelling-overview}). We report these methods by TCDP rule family.

\definecolor{tabblue}{RGB}{55,115,216}
\newcommand{\cic}[4]{\cellcolor{tabblue!#1}${#2}^{+#3}_{-#4}$}

\begin{wraptable}{r}{0.53\textwidth}
  \vspace{-\baselineskip}
  \centering
  \small
  \caption{Accuracy on NBA under the four timestamp conditions. Super- and subscripts indicate the 95\% bootstrap interval over questions.}
  \label{tab:nba150-accuracy}
  {\footnotesize\setlength{\tabcolsep}{2pt}\renewcommand{\arraystretch}{1.25}%
  \begin{tabularx}{\linewidth}{l*{4}{>{\centering\arraybackslash}X}}
    \toprule
     & \shortstack{GLM-5.3-\\Flash} & \shortstack{Sonnet-\\5} & \shortstack{Qwen3.6-\\35B-A3B} & \shortstack{Qwen3.5-\\9B} \\
    \midrule
    FULL     & \cic{67}{0.973}{.020}{.027} & \cic{69}{0.980}{.020}{.027} & \cic{54}{0.913}{.040}{.047} & \cic{5}{0.680}{.073}{.073} \\
    RELATIVE & \cic{70}{0.987}{.013}{.020} & \cic{66}{0.967}{.027}{.033} & \cic{60}{0.940}{.033}{.040} & \cic{5}{0.680}{.073}{.073} \\
    NO-TS    & \cic{62}{0.947}{.033}{.040} & \cic{62}{0.947}{.033}{.040} & \cic{43}{0.860}{.053}{.060} & \cic{5}{0.513}{.080}{.080} \\
    SHUFFLED & \cic{45}{0.867}{.053}{.057} & \cic{49}{0.887}{.047}{.053} & \cic{39}{0.840}{.060}{.060} & \cic{5}{0.593}{.080}{.080} \\
    \bottomrule
  \end{tabularx}}
  \vspace{-\baselineskip}
\end{wraptable}

\newpage
\subsection{NBA: Pre-Alignment Stands In for Timestamps}

\textbf{Accuracy.} For the three stronger models, \textsc{Relative} keeps accuracy within the \textsc{Full} interval, \textsc{No-TS} lowers it slightly, and \textsc{Shuffled} lowers it most (Table~\ref{tab:nba150-accuracy}). Removing clock values therefore matters little, while breaking alignment matters most. Qwen3.5-9B is the exception, with a larger drop under \textsc{No-TS} than under \textsc{Shuffled}.


\textbf{Trajectories.} GLM-5.3-Flash answers through  shortcuts that need no clock (Appendix Figure~\ref{fig:nba-methods}). 
\textit{Temporal causal structure:} no question requires the time between plays, so this structure is not tested.
\textit{Time semantics:} 61--66\% read the score change at the marked row and 21--25\% rule out options with basketball knowledge; neither uses time. \textit{Alignment:} the marked row indicates where the missing event belongs, so the model never aligns text and numbers; under \textsc{Shuffled}, only 1\% notice the mismatch.
\textbf{The benchmark does the alignment for the model, so it rarely tests whether timestamps are used.}

\subsection{GDELT: Testing What Order Hides}

\begin{wrapfigure}{r}{0.45\textwidth}
  \vspace{-\baselineskip}
  \centering
  \includegraphics[width=\linewidth]{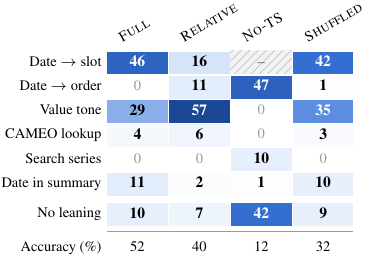}
    \caption{Decisive methods of GLM-5.3-Flash on GDELT (\% of runs). \emph{Date$\to$slot}: puts a recalled date on a listed time; \emph{Date$\to$order}: orders events by recalled dates; \emph{Value tone}: matches story tone to GoldStein values; others in Appendix~\ref{app:gdelt-categories}. Hatched: impossible in that condition. }

  \label{fig:accuracy}
  \vspace{-\baselineskip}
\end{wrapfigure}

\textbf{The control.} GDELT lists the events in random order, so, unlike on NBA, order cannot replace the timestamps. Without them (\textsc{No-TS}), GLM-5.3-Flash's accuracy falls from 52\% to 12\%, and many runs never pick an option (Figure~\ref{fig:accuracy}; Sonnet-5 in Appendix~\ref{app:gdelt}). For this model, timestamps are needed.

\textbf{Time semantics.} 
\textsc{Relative} keeps the series shape but removes the dates. Fewer runs then match a remembered story to a listed date (\emph{Date$\to$slot}, 46\% to 16\%), and more match each story's tone to the series values (\emph{Value tone}, 29\% to 57\%), a correlational shortcut. Accuracy falls from 52\% to 40\%, so the model relies on the dates themselves, not just their spacing (Figure~\ref{fig:accuracy}).


\textbf{Alignment.} \textsc{Shuffled} keeps real dates but attaches them to the wrong events. The model keeps its methods, now applied to the wrong times, and accuracy falls from 52\% to 32\%. Only about a third of runs notice the mismatch (Appendix~\ref{app:agreement}): the model uses timestamps but rarely checks them, violating TCDP's conflict-detection rule (3d in Appendix Table~\ref{tab:tcdp-audit}).

\textbf{What substitutes for timestamps.} Across both tasks, models answer through shortcuts that replace timestamps.
On GDELT, they are story tone once dates are removed, and likely memory of the events, which keeps Sonnet-5 at 43\% without any timestamps (chance is 25\%).
Because these shortcuts are enough to answer the questions, the two tasks together make only three of TCDP's ten rules necessary (Appendix Table~\ref{tab:tcdp-audit}).

\section{Conclusion}
\label{sec:conclusion}

We framed timestamps as \emph{world indices} and tested whether LLMs use them by altering timestamps in two TSRBench tasks and auditing each reasoning trajectory with TCDP. On NBA, event order and the marked row replace the timestamps, so removing clock values barely changes the stronger models' accuracy. On GDELT, where order is withheld, models use dates to recall world events, yet most runs miss wrong times and switch to tone matching when dates are removed. Accuracy alone therefore overstates temporal reasoning, and benchmarks should also check how answers are reached.


\bibliographystyle{plainnat}
\bibliography{reference_neurips}


\newpage
\appendix

\section{Related Works}

\textbf{Time Series Reasoning.}
Adapting LLMs to address time series tasks begins with serializing values into prompts for zero-shot inference~\citep{xue2023promptcast, gruver2023large, liu2024lstprompt}, which proves brittle for temporal inference~\citep{merrill2024language, tan2024language}. Subsequent work aligns the two modalities by fine-tuning on TS--text corpora~\citep{xie2024chatts, kong2025time, wang2025chattime, langer2025opentslm, yu2025ts}, and more recently elicits long chain-of-thought through reinforcement learning~\citep{niu2025langtime, zhang2025timemaster, guan2026timeomni}.
Benchmarks have also evolved in parallel. TimeSeriesExam~\citep{cai2024timeseriesexam} and Time-MQA~\citep{kong2025time} evaluate understanding over isolated numerical sequences, stripping away the context in which the series was generated. Time-MMD~\citep{liu2024time} and CiK~\citep{williams2024context} reintroduce context, but only as background for forecasting rather than as evidence to be reasoned over. MTBench~\citep{chen2025mtbench} and Time-IMM~\citep{chang2026time} pair news and events with series to probe temporal reasoning, yet confine it to narrow domains, while TSRBench~\citep{yu2026tsrbench} broadens the task space across perception, reasoning, prediction, and decision-making. In all of these benchmarks, the event series are already aligned, so LLMs inherit the very correspondence that temporal reasoning is supposed to establish; their use of timestamps is never actually tested.

\textbf{The Role of Timestamp in Time Series.}
In numerical forecasting, timestamps enter the model as calendar covariates: Informer~\citep{zhou2021informer} and Autoformer~\citep{wu2021autoformer} sum them into the value embedding, while DLinear~\citep{zeng2023transformers} discards them altogether without loss. Other work captures temporal variation directly from the numerical observations, through adaptive normalization~\citep{ye2024frequency}, or test-time adaptation~\citep{deng2026correction}. GLAFF~\citep{wang2024rethinking} and CFPT~\citep{kou2025cfpt} revisit this design and show that timestamps carry global periodic guidance that local observations cannot supply, yet both exploit them as a seasonality prior rather than as a reference to the world. A second line treats timestamps as metric spacing: continuous-time encodings~\citep{shukla2021multi, zhang2025unleashing} recover the irregular gaps that fixed-step tokenization erases, but the information they restore remains internal to the series. On the textual side, timestamps are instead a key into world knowledge~\citep{dhingra2022time}, and benchmarks such as TempReason~\citep{tan2023towards}, TRAM~\citep{wang2024tram}, TimeBench~\citep{chu2024timebench}, and Test of Time~\citep{fatemi2025test} show that LLMs often answer time-sensitive questions from memorized facts or from the order in which facts are presented~\citep{chen2024premise} rather than from the dates themselves. Each line is partial, reading timestamps either as structure internal to the series or as pointers to a world outside it, and neither treats them as the index binding an observation to its generating event, which is precisely the role we study.


\section{Implementation Details}
\label{app:implementation}

\subsection{Tasks}
\label{app:tasks}

\paragraph{NBA.}
Each of the 150 questions comes from one game.
The model sees the play-by-play events before and after one missing event, in game order, and a numerical table with the score of each team and their win probabilities over time.
The table marks the row of the missing event.
The model chooses which of four candidate events filled the gap.

\paragraph{GDELT.}
Each of the 160 questions gives 6--8 news summaries, a GoldSteinScale series and a list of event times.
GoldSteinScale scores each recorded event from $-10$ (conflict) to $+10$ (cooperation).
The event times are listed in random order, and the prompt does not say which time belongs to which event.
The model chooses the true order of the events from four options.
Ordering the events is the same as giving each event one of the listed times, and so the series value at that time.

\subsection{Data Interventions}
\label{app:interventions}

Each intervention changes only the timestamps.
The events, the numerical values, the question, the four options and the gold answer stay the same in every condition.
\textsc{Full} is the original input.
The other three conditions remove one kind of information that a timestamp carries.
\textsc{Relative} gives each time as an offset from the target (NBA, seconds) or first record (GDELT, hours), keeping order and spacing but losing game or calendar time.
\textsc{No-TS} removes all timestamps.
\textsc{Shuffled} keeps real times but breaks the alignment between events and series.

\begin{figure*}[t]
  \centering
  \includegraphics[width=\textwidth]{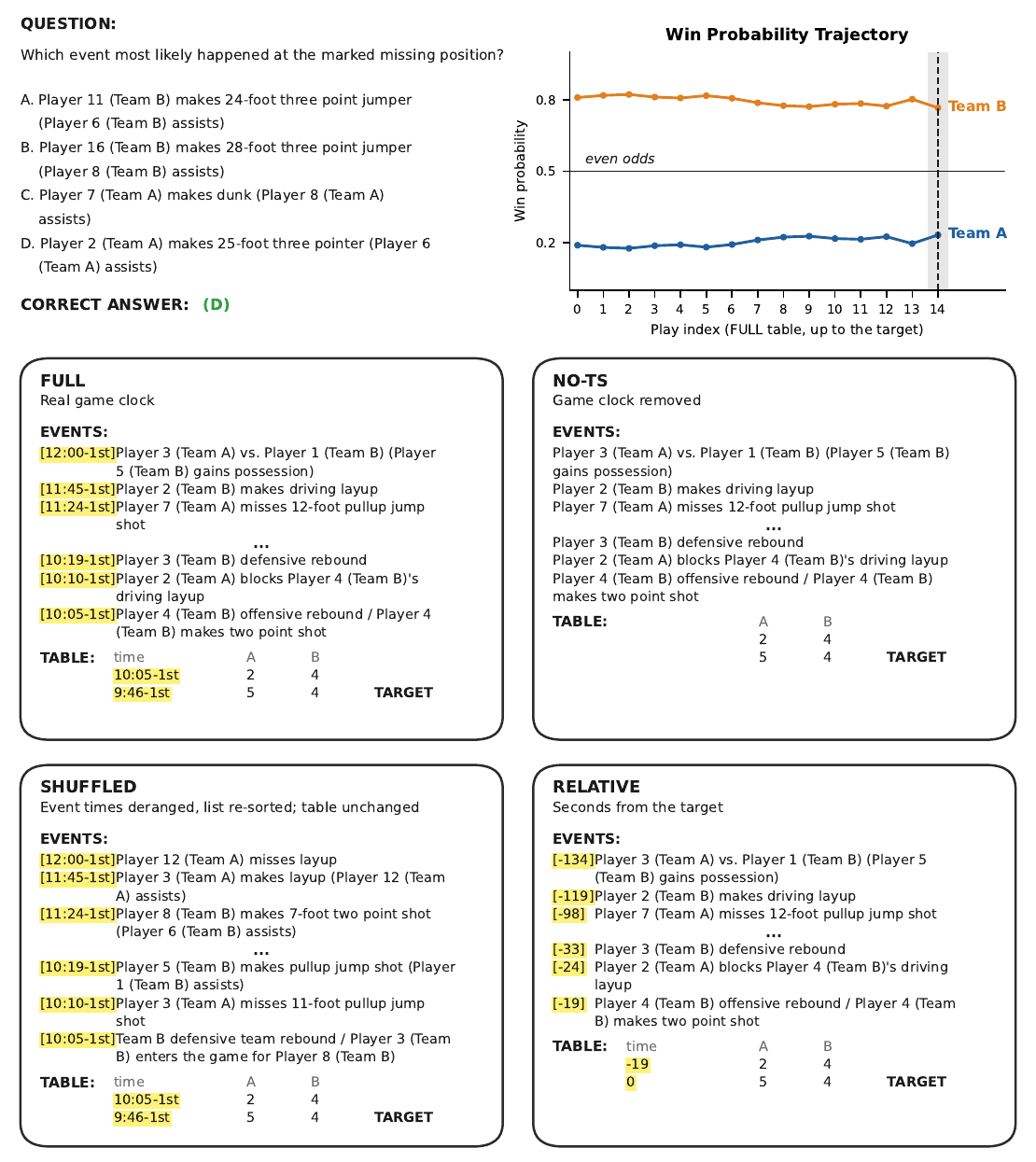}
  \caption{
  \textbf{Data interventions on an NBA question.}
  The events, numerical values, question and options are the same in every condition.
  \textsc{Full} keeps the game clock;
  \textsc{Relative} gives seconds from the target;
  \textsc{No-TS} removes the clock;
  \textsc{Shuffled} gives each event the time of another event and re-sorts the list, while the numerical table keeps its real clock.
  }
  \label{fig:data_intervention}
\end{figure*}

\paragraph{NBA.}
Figure~\ref{fig:data_intervention} shows one question under the four conditions.
\begin{itemize}
  \item \textsc{Relative} converts every time, in the event list and in the numerical table, to the exact number of seconds from the target.
  The target is at 0, earlier rows are negative and later rows are positive.
  Intervals and the event--row correspondence are kept, but the game clock and the quarter are lost.
  \item \textsc{No-TS} deletes the game clock from the event list, the numerical table and the target moment.
  The listed order, the split into past and future events, and the target marker remain.
  \item \textsc{Shuffled} changes only the event list.
  Each observed event receives the time of another event, through a seeded derangement, so no event keeps its own time.
  The events are then sorted by their new times, so the list still runs forward.
  The numerical table, its real clocks, the target marker and the target moment are identical to \textsc{Full}.
  As a result, the text and the table no longer agree on when each event happened.
\end{itemize}

\begin{figure*}[t]
  \centering
  \includegraphics[width=\textwidth]{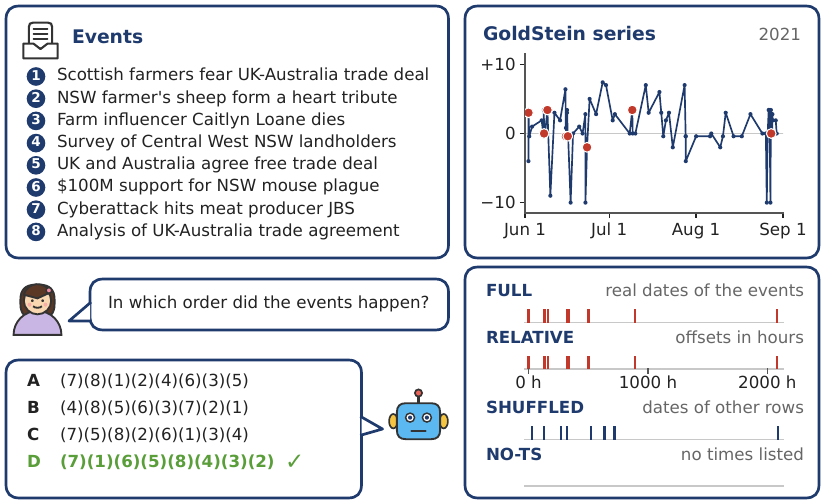}
  \caption{
  \textbf{Data interventions on a GDELT question.}
  Left: short summaries of the eight news events and the four options.
  Right: the GoldStein series of the question; red dots mark the moments of the events.
  The strips below mark where the listed event times fall on the same time axis.
  \textsc{Relative} lists the same moments as \textsc{Full}, written as hours since the first record;
  \textsc{Shuffled} lists other moments of the series;
  \textsc{No-TS} lists no times.
  }
  \label{fig:gdelt-example}
\end{figure*}

\paragraph{GDELT.}
Figure~\ref{fig:gdelt-example} shows one question under the four conditions, and Table~\ref{tab:gdelt-example} gives its listed times as the prompts show them.
\begin{itemize}
  \item \textsc{Relative} converts every time, in the series and in the list of event times, to hours since the first row of the series.
  Intervals and the event--row correspondence are kept, but the calendar dates are lost.
  \item \textsc{No-TS} replaces the series timestamps with row positions and deletes the list of event times.
  \item \textsc{Shuffled} leaves the series byte-identical to \textsc{Full}.
  Each listed event time is replaced by another timestamp of the same series, through a derangement of the series timestamps.
  Every listed time is therefore a real row of the series, but none is the moment of an event.
\end{itemize}
In every condition the event times are listed in random order, and the prompt does not say which time belongs to which event.

\begin{table*}[t]
  \centering
  \caption{
  \textbf{Data interventions on a GDELT question} (eight events, gold order (7)(1)(6)(5)(8)(4)(3)(2)).
  The first column gives the index of the series rows; the second gives the list of event times exactly as the prompt shows it, in random order.
  Dates are in 2021.
  }
  \label{tab:gdelt-example}
  {\footnotesize\renewcommand{\arraystretch}{1.25}%
  \begin{tabular}{lll}
    \toprule
    Condition & Series index & Listed event times \\
    \midrule
    \textsc{Full}     & date and time        & 07-09, 06-16, 06-07, 06-09, 06-15, 06-23, 08-27, 06-02 \\
    \textsc{Relative} & hours since first row & 888.5, 337, 133.5, 162.5, 320.5, 501.5, 2077.75, 0.25 \\
    \textsc{No-TS}    & row position         & (no list) \\
    \textsc{Shuffled} & as \textsc{Full}     & 06-13, 08-28, 07-02, 06-28, 06-07, 06-24, 06-03, 06-15 \\
    \bottomrule
  \end{tabular}}
\end{table*}

\subsection{Models and Scoring}
\label{app:models}

On NBA we evaluate \texttt{GLM-5.3-Flash}, \texttt{Sonnet-5}, \texttt{Qwen3.6-35B-A3B} and \texttt{Qwen3.5-9B}; on GDELT, \texttt{GLM-5.3-Flash} and \texttt{Sonnet-5}.
\texttt{GLM-5.3-Flash} and both Qwen models run with thinking enabled, temperature 0, top-$p$ 1 and a fixed seed, and at most 22{,}000 completion tokens.
\texttt{Sonnet-5} runs through the Claude Code command line with all tools disabled and a fresh session per question; its sampling temperature is not set by us.
Every question is asked once in each condition.
The system prompt asks the model to use only the information in the prompt and no external information.
A response counts as correct only if it gives a valid answer that matches the gold answer.
A response without a valid answer, including a run that hits the token limit, counts as wrong.
Intervals are 95\% bootstrap intervals over questions.

\section{Trajectory Labelling}
\label{app:reasoning-procedures}

\subsection{Overview}
\label{app:labelling-overview}

We label every \texttt{GLM-5.3-Flash} run in the four conditions with the method that decides its answer.
Each label comes from a language model that reads the full reasoning trace: Claude Opus~5 for NBA and Claude Opus~5.5 for GDELT, since Opus~5 was no longer available when the GDELT analysis was run.
The labeller is not told whether the run was correct.

\begin{figure}[h]
  \centering
  \includegraphics[width=\textwidth]{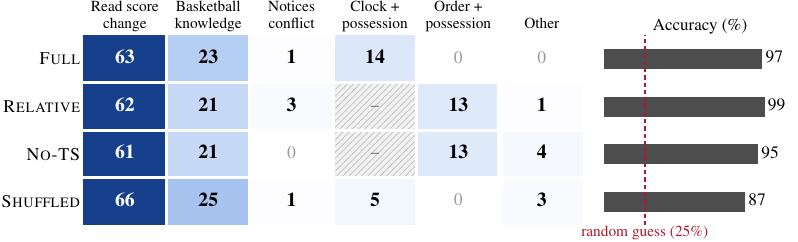}
  \caption{Decisive methods of GLM-5.3-Flash on NBA, as a share of the 150 runs in each condition (P3 labels, \%). Hatched cells mark a clock-based method, which cannot occur when absolute clock values are absent; 0 marks a method that can occur but no run uses. Other includes the runs without a final answer. The right panel gives accuracy over all 150 questions (Table~\ref{tab:nba150-accuracy}); the dashed line marks a random guess among four options (25\%).}
  \label{fig:nba-methods}
\end{figure}

\paragraph{NBA.}
We label the 591 runs that give a final answer (600 runs, less 9 that hit the token limit).
The final labels come from one condition-blind pass (P3 below): each trace is shown under an opaque id, without its condition.
The runs without an answer are counted under \emph{Other} in Figure~\ref{fig:nba-methods}.

\paragraph{GDELT.}
We label all 632 runs (\textsc{Full} 158, \textsc{Relative} 160, \textsc{No-TS} 154, \textsc{Shuffled} 160). The remaining 8 of the 640 runs (2 in \textsc{Full}, 6 in \textsc{No-TS}) failed with API connection errors in the first collection and were re-run after the labelling; to keep the labelled and scored runs identical, Table~\ref{tab:gdelt-accuracy} uses the original failures, which count as wrong.
A run that hits the token limit is given the method behind the answer it was leaning towards, and counts as wrong.
A run with no leaning is shown as \emph{No leaning}.
The labeller sees the condition of each run.
Whether a run notices that the times do not fit is therefore labelled again in a separate condition-blind pass (Section~\ref{app:gdelt-prompts}); the 36\% and 11\% in Table~\ref{tab:tcdp-audit} come from that pass.

\subsection{NBA Prompts}
\label{app:analysis-prompts}

We built the NBA taxonomy in three steps.
We first induced categories from a sample of traces (P1) and labelled every run with them (P2).
The largest first-pass categories mixed different methods, so we then labelled every run again with one prompt that gives four methods (P3).
Table~\ref{tab:nba-merge} shows how these methods relate to the earlier categories.
An intermediate split of the score-delta runs by keyword rules did not agree with a second labeller ($\kappa = 0.08$), and we do not use it.
P2 is still used in one place: it tells whether a possession chain follows the clock or the row order.
A fourth prompt (P4) labels one thing for TCDP rule 1c: whether a run states the time between two moments, and whether it uses it.
The prompts use the internal condition names \texttt{full}, \texttt{remove}, \texttt{relative} and \texttt{shuffle}, for \textsc{Full}, \textsc{No-TS}, \textsc{Relative} and \textsc{Shuffled}.

\paragraph{Stage 1: Taxonomy induction (P1).}

\begin{promptbox}
You are reviewing summaries of the reasoning strategy glm-5.3-flash used
across 80 NBA play-by-play reasoning runs (same 20 questions under four
timestamp representations: full, relative, remove, shuffle).

Propose 5-8 mutually exclusive, collectively exhaustive categories that
capture the distinct reasoning strategies you observe. Each category
should have:

- a short snake_case name
  (e.g. "absolute_timestamp_anchoring", "score_delta_inference")
- a one-sentence description of what the strategy looks like

Prioritize categories that distinguish HOW the model uses (or ignores)
temporal / ordering information, since that is the experimental variable.
\end{promptbox}

\paragraph{Stage 2: First-pass classification (P2).}

\begin{promptbox}
This is one glm-5.3-flash run answering an NBA play-by-play reasoning
question. The timestamp condition for this run is:
(full = original clock times; relative = times relative to the target
moment; remove = timestamps stripped out; shuffle = timestamps randomly
scrambled).

Classify the model's PRIMARY reasoning strategy using exactly one of
these categories:

- absolute_clock_score_delta:
  The model anchors adjacent states to game-clock timestamps and treats
  the resulting score change or non-change as the primary determinant
  of the missing event.

- absolute_clock_possession_chain:
  The model uses absolute clock chronology to reconstruct possession
  transfers among fouls, misses, rebounds, turnovers, and later plays,
  with score state serving mainly as a constraint.

- relative_order_score_delta:
  The model ignores or lacks usable game-clock times and infers the
  event chiefly from score movement between the pre-target, target,
  and post-target positions.

- relative_order_possession_chain:
  The model relies on before/after or relative timestamps to build a
  causal possession sequence, typically using a following rebound or
  shot to infer the target event.

- win_probability_proxy:
  When the score is unchanged or ambiguous, the model treats a
  win-probability movement as the decisive temporal signal for a miss,
  turnover, or other non-scoring event.

- noisy_timeline_local_override:
  The model identifies timestamp, score, or event-order misalignment
  and deliberately overrides the global chronology with the most
  coherent local score, rebound, or possession cue.

- lineup_roster_elimination:
  When temporal and score evidence do not uniquely resolve the target,
  the model primarily selects by substitutions, on-court status,
  roster validity, team consistency, or recurring player combinations.

- event_plausibility_tiebreak:
  After temporal evidence narrows the possibilities, the model resolves
  the remaining tie using basketball conventions or event plausibility,
  such as normal shot distance, rebound semantics, or required
  follow-up actions.

Also judge how the model handled the play-by-play TIMESTAMPS / event
ordering:

- trusted:
  it used the timestamps or their implied ordering as a reliable basis
  for its answer

- flagged_unreliable:
  it noticed the timestamps/ordering were inconsistent or untrustworthy
  and worked around them

- ignored:
  timestamps were present but it did not rely on them

- absent:
  there were no timestamps in the input to use

Base both judgments on the model's actual reasoning, and cite the evidence.
\end{promptbox}
The full pass used the same categories with added decision rules and a fixed output format.

\paragraph{Stage 3: Final labelling (P3).}
\texttt{\{\{RUN\}\}} is replaced by the list of trace files of one batch and the output path.
Each trace file holds an opaque run id, the four options, the reasoning and the final output.

\begin{promptbox}
You are labelling reasoning traces written by GLM-5.3-Flash on an NBA
play-by-play task. Each question shows the play-by-play events before and after
one missing event, a numerical table of each team's score and win probability
over time with the missing position marked, and four candidate events. The model
must choose which candidate filled the gap. The times shown to the model may be
real game clocks, offsets from the gap, absent, or wrong.

Label the method that decides the final answer. Choose exactly one:

- change_readout: the run reads the change in score (or in win probability)
  between the row before the gap and the gap row, and picks the option whose
  points or outcome match. Other checks may be mentioned, but they do not change
  the answer.
- possession_chain: the run reconstructs who has the ball across the plays
  around the gap - a following rebound, turnover, foul or inbound - and picks
  the option that this sequence requires.
- knowledge_screen: the run rules out options with knowledge the numbers do not
  give: scoring rules (and-one, free-throw counts, technical or flagrant fouls),
  who is on the court, which team a player is on, whether the play is plausible,
  or whether it fits the event text.
- conflict_noticed: the run states that the times, scores or events do not line
  up - for example that the clock is scrambled, or that the table and the text
  disagree - and this shapes the answer: it falls back on the row order,
  overrides with a local cue, or answers while leaving the conflict open.
- other: none of the above decides the answer.

Decision rules:
1. Choose by what decides the final answer, not by what is mentioned.
2. A passing remark that something looks odd is not conflict_noticed if the run
   then answers from the score change as if nothing were wrong.
3. If two methods each narrow the options, choose the one that makes the final
   choice.

{{RUN}}

Write one JSON object per run, one per line, to the output file - nothing else in
that file. Use exactly this shape:

{"run": "<run id from the file header>", "method": "...", "evidence": "<one sentence quoting the reasoning>", "confidence": 0.0}
\end{promptbox}

\paragraph{Lag check (P4).}
\texttt{\{\{RUN\}\}} is replaced as in P3; P4 reads the same trace files under the same opaque ids, so the labeller does not see the condition.
Table~\ref{tab:nba-lag} gives the result.
No question needs a lag, and under \textsc{Full} few runs compute one.
Under \textsc{Shuffled} more runs compute a gap and use it; these gaps come from the listed times, which \textsc{Shuffled} has moved.

\begin{table}[h]
  \centering
  \caption{NBA runs that state the time between two moments (computed) and use it to keep or rule out an option (used). P4 labels, \% of the 150 runs in each condition.}
  \label{tab:nba-lag}
  {\footnotesize\renewcommand{\arraystretch}{1.2}%
  \begin{tabular}{lcccc}
    \toprule
     & \textsc{Full} & \textsc{Relative} & \textsc{No-TS} & \textsc{Shuffled} \\
    \midrule
    Lag computed & 13 & 17 & 0 & 27 \\
    Lag used     & 2  & 3  & 0 & 13 \\
    \bottomrule
  \end{tabular}}
\end{table}

\begin{promptbox}
You are labelling reasoning traces written by GLM-5.3-Flash on an NBA
play-by-play task. Each question shows the play-by-play events before and after
one missing event, a numerical table of each team's score and win probability
over time with the missing position marked, and four candidate events. The
model must choose which candidate filled the gap. The times shown to the model
may be real game clocks, offsets in seconds from the gap, absent, or wrong.

You label one thing only: whether the run works out how much time passed
between two moments, and whether it uses that amount.

- lag_computed (true/false): true if the run states an amount of elapsed time
  between two moments, worked out from the times in its prompt - for example
  "from 9:52 to 9:47 is 5 seconds" or "only 3 seconds later". Naming two times,
  or saying that one play came before or after another, is false: order alone
  is not a lag.
- lag_used (true/false): true if that elapsed time is used as a limit on what
  could have happened in the gap - for example that there was too little time
  for a possession change or a second shot - and it keeps or rules out at least
  one option. If lag_computed is false, lag_used is false.
- lag_quote: if lag_computed, the sentence (at most 200 characters) where the
  run states the elapsed time; otherwise "".
- confidence: 0 to 1, how sure you are of the two booleans.

Rules:
1. Label what the run wrote, not whether it was right.
2. Copy every quote exactly from the trace: same characters, same spacing, no
   ellipses.

{{RUN}}
Write one JSON object per run, one per line, to the output file - nothing else
in that file. Use exactly this shape:
{"run": "<run id from the file header>", "lag_computed": false, "lag_used": false, "lag_quote": "", "confidence": 0.0}
\end{promptbox}

\subsection{NBA Method Definitions}
\label{app:nba-methods}

Each run is assigned the method that decides its final answer, not every step it mentions.

\paragraph{Read score change.}
The run reads the score in the row before the gap and in the gap row, and picks the option whose points match.
Win-probability changes read the same way are included.

\paragraph{Basketball knowledge.}
The run rules out options with knowledge the numbers do not give: scoring rules, who is on the court, which team a player is on, or whether the play is plausible.

\paragraph{Notices conflict.}
The run says that the times, scores or events do not line up, and this shapes its answer: it falls back on the row order, overrides the table with a local cue, or answers while leaving the conflict open.

\paragraph{Clock + possession.}
The run follows who has the ball across the plays around the gap, using the game clock to place them.

\paragraph{Order + possession.}
The same possession chain, built from the row order or the offsets instead of the clock.

\paragraph{Other.}
No method above decides the answer, or the run gives no final answer.

\begin{table}[h]
  \centering
  \caption{How the final NBA methods relate to the earlier categories (P2 and the keyword refinement).}
  \label{tab:nba-merge}
  {\footnotesize\renewcommand{\arraystretch}{1.2}%
  \begin{tabular}{ll}
    \toprule
    Final method & Earlier categories \\
    \midrule
    Read score change    & score only; win-probability proxy \\
    Basketball knowledge & basketball rules; option check; lineup/roster elimination; plausibility tiebreak; game-phase cue \\
    Notices conflict     & reject bad clock; unresolved conflict; noisy-timeline override \\
    Clock / Order + possession & clock + possession; order + possession; replay events \\
    \bottomrule
  \end{tabular}}
\end{table}

\subsection{GDELT Prompts}
\label{app:gdelt-prompts}

We induced the GDELT categories with Docent~\citep{meng2025docent} from 120 traces in two steps.
The first step writes a short account of each run's decisive method (G1); the second proposes categories from all accounts (G2).
We then labelled every run with G3.
\textsc{No-TS} keeps its labels from an earlier version of G3.
The later versions change only the descriptions of the rebuilt conditions, the timestamp-handling field and a rule for \textsc{Relative}; none of these applies to a run with no listed times, and the eight categories are the same.
G4 labels only whether a run notices that the listed times or the values at them do not fit, and whether it uses them.
It hides the condition and mixes runs of \textsc{Full}, \textsc{Relative} and \textsc{Shuffled} in one order.

\paragraph{Induction, step 1 (G1).}

\begin{promptbox}
Write a short account of this run's DECISIVE METHOD, in two labelled parts.

METHOD: the single line of reasoning that decided the run.

  If the run answered: find the step the final letter actually came out of.
  Traces often try several approaches and abandon most of them; name only the
  one that produced the choice, even if it appears late or briefly. Say
  concretely what information it used (which part of the prompt, or what
  outside the prompt) and how that information was turned into a choice among
  A-D - for example by fixing some events' positions and eliminating options,
  by matching something to the series values, or by ordering events on some
  other basis.

  If the run did not answer: say so in the first clause. Then name the method
  it was pursuing in its final stretch, in the same concrete terms, and what it
  was still trying to resolve when it was cut off. Do not guess which letter it
  would have chosen.

OTHERS: in one sentence, the main approaches the trace tried and dropped before
or around the decisive one. These are context only.

Describe what the model did, not whether it was right. You have not been told
whether this run was correct and should not try to work it out. Name the
temporal material each step used - calendar dates, the deranged dates, relative
indices, bare series positions, the series values, or nothing temporal - since
the conditions differ in which exist. Keep the account under 130 words and cite
the parts of the trace you describe, especially the decisive step.
\end{promptbox}

\paragraph{Induction, step 2 (G2).}

\begin{promptbox}
Propose between five and nine categories of DECISIVE METHOD that together cover
these runs.

WHAT THEY ARE FOR

Every one of 612 runs will be assigned to exactly one category. The result is a
table of method by condition (FULL, SHUFFLED, RELATIVE, NO_TS), reported
separately for runs that answered and runs that did not. Its question is how
the way the model decides changes when the timestamps are deranged, reduced to
indices, or removed.

REQUIREMENTS

- Mutually exclusive and jointly exhaustive: a reader holding one run must be
  able to put it in exactly one category. Where a run combines methods, the
  category is the one the choice (or the final stretch) actually rests on; say
  in each description how to decide that against its nearest neighbour.
- About method, not outcome: what information the decision rests on and how it
  turns that into a choice among A-D. Not whether it was right, and not the
  news content.
- Usable for both kinds of run: each category must be assignable to a run that
  answered (the method its answer came from) and to one that did not (the
  method it was pursuing when cut off).
- Phrase categories around what the model does with whichever temporal marker
  it has where you can, so the same category can occur in several conditions.
  Where a method is only possible in some conditions - it needs calendar dates,
  say - keep it if it is real, but say which conditions it can occur in, so
  that a zero elsewhere is read as impossible rather than avoided.
- Using outside knowledge of real-world dates, or reverse-mapping series
  values through a remembered CAMEO table, are methods like any other here and
  may be categories in their own right if runs decide on them.
- Include a category for runs that end by falling back to a guess or an
  arbitrary pick if such runs exist; do not include a catch-all "other".
- No category should hold nearly every run. If most runs share one broad
  method, split it by what distinguishes them.

For each category give:
- name: short, snake_case
- description: what a run in it does, concretely enough that two readers would
  assign the same run the same way
- decision_rule: one sentence a reader applies to a run to decide membership
- nearest_neighbour: the category it is most easily confused with, and how to
  tell them apart
- possible_in: the conditions in which this method can occur at all
\end{promptbox}

The run count in G2 (612) reflects the runs available when the categories were induced; the remaining 20 had failed with API connection errors and were re-run afterwards.

\paragraph{Labelling (G3).}
\texttt{\{\{RUN\}\}} is replaced by the trace files of one batch.

\begin{promptbox}
You are labelling reasoning traces written by GLM-5.3-Flash on a multiple-choice
benchmark. Read the background first; it contains facts about the task that the
model itself was not told.

BACKGROUND

Each run is one attempt at a question built from TSRBench. The model is given 6
to 8 GDELT news-event summaries, labelled (1), (2), (3) and so on, and asked to
put them in the order they actually happened, choosing between four candidate
orderings (A-D). Depending on the condition it is also given a GoldSteinScale
time series and a list of event times.

Facts about how these questions are built:

1. The list of event times is explicitly shuffled and unlabelled: the prompt
   says the times are "presented in a random order" and never says which time
   belongs to which event. Recovering that pairing is the core difficulty.
2. Position carries no information. Assuming the k-th listed time belongs to
   event (k) is wrong by construction.
3. GoldSteinScale is a fixed lookup from the CAMEO code of the action reported,
   not a measure of how good or bad the news was. Two candidate times often
   share a value.
4. The system prompt forbids using outside knowledge about when the events
   really happened.
5. The four conditions differ only in the temporal material the prompt carries:
     FULL      the timestamped series, and the event times as calendar dates;
               each listed time is the moment of one event and names a row of
               the series
     SHUFFLED  the series exactly as in FULL, but every listed event time has
               been replaced by the timestamp of a different series row, on
               average about two weeks away. The listed times still name rows
               of the series, but none of them is the moment of its event, so
               the values at those rows do not belong to the events either
     RELATIVE  every timestamp, in the series and in the list, replaced by the
               number of hours since the series' first record. Order, gaps and
               which row each event time names all survive; only the calendar
               anchor is gone
     NO_TS     every timestamp deleted, and the list of event times removed;
               the series is bare positions and values

About the traces:

A. Many runs never produced an answer: generation was capped at 22,000
   completion tokens, and those traces end mid-thought. You are told which kind
   each run is.
B. Text that was pure repetition - a passage pasted again word for word, or a
   long run of filler such as "Hmm, hmm." - has been removed, and a marker in
   square brackets says how many characters were removed. Everything else is
   the model's text as written. Do not treat a marker as something the model
   wrote; do treat it as evidence that the model was repeating itself there.

{{RUN}}

WHAT TO LABEL

Label what the model did, not whether it was right. You have not been told the
correct ordering and should not try to work it out.

---------------------------------------------------------------------------
1. decisive_method - exactly one of the eight categories below
---------------------------------------------------------------------------

A long trace raises and drops many approaches. Label only the one that decided
the run:

- If the run answered: the line of reasoning the final letter actually came out
  of. Find the step at which the model commits to its letter and ask what that
  step rests on. It may appear late or briefly; what the trace spends most of
  its length on is not the criterion.
- If the run did not answer: the method the model was actively pursuing in its
  final stretch, before it was cut off. If the final stretch is itself
  repetition, label the method being repeated.

The categories:

external_calendar_slot_matching
  The run recalls or guesses real-world publication or event dates from outside
  the prompt, sorts the supplied calendar timestamps, and assigns one or more
  events to those dated slots; the resulting slot positions eliminate options.
  Rule: the decisive outside-world date matters specifically because it is
  paired with one of the prompt's calendar timestamps.
  Nearest neighbour: external_direct_chronology - choose that instead when
  recalled dates establish only an event-to-event order and the supplied
  calendar slots are ignored or unnecessary.

external_direct_chronology
  The run reconstructs the stories' actual chronology from remembered
  real-world dates, historical facts or publication timing, and directly
  matches that event sequence or pairwise order to A-D. It may ignore temporal
  material the prompt does carry, or compensate for its absence with an outside
  timeline.
  Rule: remembered external chronology itself selects or narrows the option,
  without needing to map events onto the prompt's calendar-date slots.
  Nearest neighbour: intrinsic_text_temporal_constraints - external imports a
  date or sequence not stated in the summaries; intrinsic derives its ordering
  from wording or dates contained in the prompt.

intrinsic_text_temporal_constraints
  The run extracts temporal constraints from the summaries themselves -
  explicit dates, scheduled future actions, retrospective periods, narrative
  stages, before/after language - and uses them either to match supplied
  markers or to eliminate options directly.
  Rule: the decisive ordering constraint is recoverable from the prompt's event
  text without recalling an unstated real-world date.
  Nearest neighbour: external_direct_chronology - if the model must remember
  when the named event or article happened, rather than reason from text in the
  summary, it is external.

cameo_lookup_matching
  The run tries to identify each summary's CAMEO action code or exact fixed
  GoldStein weight, often reconstructing a remembered codebook, and maps those
  action-specific values to temporal markers or to the assignments the options
  imply.
  Rule: the active inference depends on an exact or action-specific
  CAMEO-to-GoldStein lookup. Semantic_value_matching instead judges stories as
  positive, negative or neutral.
  Nearest neighbour: semantic_value_matching - which treats the values as
  general sentiment, cooperation, conflict or intensity without committing to a
  specific action weight.

semantic_value_matching
  The run treats GoldStein values as broad semantic signals - positive versus
  negative, cooperative versus conflictual, neutral versus intense - and
  selects the option whose event-to-value assignments seem most plausible. It
  may anchor on one distinctive value or compare whole option mappings.
  Rule: broad tone or event-type plausibility against the values is the
  decisive basis, and neither an exact CAMEO lookup nor a search through
  repeated occurrences in the wider series is essential.
  Nearest neighbour: cameo_lookup_matching - coarse polarity or intensity here,
  the fixed action-code value there.

series_occurrence_search
  The run first infers likely values for events, then searches the wider time
  series for occurrences of those values, chooses particular positions or a
  nearby or increasing cluster, and orders events by the positions it chose.
  Rule: selecting occurrences or subsequences within the wider series - not
  merely scoring the supplied event-time slots - is necessary to produce the
  ordering.
  Nearest neighbour: semantic_value_matching - both infer values from content,
  but this one also chooses among repeated positions in the series and derives
  the order from them.

positional_alignment_assumption
  The run assumes a correspondence created by presentation position - that
  event (k) receives the k-th listed time, that a list of indices encodes event
  numbers by position, or that event (n) sits at series position n - and sorts
  that assumed mapping.
  Rule: the final active method obtains the event-time pairing from list order,
  event number, or positional identity. Intrinsic_text_temporal_constraints
  obtains the pairing from event content and temporal evidence.
  Nearest neighbour: intrinsic_text_temporal_constraints - which derives the
  correspondence from dates or sequencing in the summaries, not from layout.

unsupported_fallback_pick
  The run ends by guessing, choosing arbitrarily among unresolved options,
  following a superficial pattern in the options, or naming a letter with no
  remaining constraint that favours it.
  Rule: no date, text constraint, value interpretation, series search or
  positional correspondence actually supports the final pick or the final
  active direction.
  Nearest neighbour: semantic_value_matching - an explicitly weak "least
  implausible" comparison still belongs there if score-content fit favours one
  option.

Rules for applying them:

1. Decide by the step the letter came out of (or the final stretch), not by
   what the model mentions first, most often or most confidently.
2. When several methods feed one decision, choose the one that supplies the
   constraint which eliminates the last rival options. A method that only
   confirms an already-chosen letter, or that the model set aside, is context.
3. A date written in a summary is intrinsic; a date the model brings from
   memory is external, even when the model presents it as obvious.
4. Between the two external categories, ask whether the recalled date was
   placed onto one of the prompt's listed times. Placed: slot matching. Used
   only to order events against each other: direct chronology.
5. A trace that merely notes the list of times is in random order, or that two
   values tie, has not thereby chosen a method; keep reading to see what it did
   about it.
6. external_calendar_slot_matching needs dated slots. In FULL and SHUFFLED the
   listed times are calendar dates. In RELATIVE it counts only if the run
   explicitly anchors a date - recalled, or written in a summary - to one of
   the hour values, converts the others to dates (or offsets from that date),
   and the deciding recalled date is then placed on one of those converted
   slots. A RELATIVE run that only orders recalled dates against each other,
   or converts the hours only to confirm a choice already made on other
   grounds, is labelled by what decided it. NO_TS has no listed times, so a run there that
   decides on recalled dates is external_direct_chronology. The other seven
   can occur in every condition.
7. For a run that did not answer, use unsupported_fallback_pick only if its
   final stretch is explicitly about picking without a basis ("I'll just go
   with...", "pick the most likely"), not merely because it had not finished.

---------------------------------------------------------------------------
2. timestamp_handling - exactly one of five
---------------------------------------------------------------------------

How the decisive method treated the LINK BETWEEN THE LISTED EVENT TIMES AND
THE EVENTS. That link can be read two ways, and both count:
- through the series: the GoldSteinScale value at a listed time is taken as
  the value of the event that happened then;
- through the dates: a listed time is taken as the moment one of the events
  happened, for example the slot a recalled date belongs in.
This is the link the SHUFFLED condition breaks: the listed times point to the
wrong rows and are not the events' real dates.

trusted
  It relied on the link: it read series values at the listed times (or hours)
  and used them as a basis for its decision, or it placed events on the listed
  times as their real moments, without saying the times or the values there
  do not fit.
flagged_unreliable
  It stated that the link does not hold - the values at the listed times do
  not fit the events, or the listed times do not fit when the events happened
  (for example a recalled date has no matching time in the list), or the
  times or values look scrambled, inconsistent or like noise - and the
  decisive method does not rest on the part it distrusted. Whatever else it
  then used, including the other half of the link, it is flagged_unreliable.
flagged_but_used
  It stated that the link does not hold, in the same sense as above, but the
  decisive method still rests on the part it distrusted - for example it said
  the recalled dates fit no listed time, then went back to placing recalled
  dates on the listed times; or said the values look scrambled, then chose by
  the values anyway. The statement must be about the prompt's times or
  values, not a doubt about its own memory or lookup.
ignored
  Listed times were present, but the decisive method used neither the values
  at them nor the times as the events' moments, and the model never said the
  link was unreliable - for example a choice made from recalled chronology
  alone, or from the summaries' text alone.
absent
  The prompt carried no times for the values to sit at. Every NO_TS run is
  absent, whatever the model did with bare series positions.

Notes:
- The prompt itself says the list of event times is in random order, so
  remarking on that is not flagged_unreliable.
- A remark that one value is surprising, followed by using the values anyway,
  is trusted. The same holds for a remark that one recalled date is a little
  off a listed time, followed by placing the event there anyway. What
  separates flagged_but_used from trusted is a statement that the prompt's
  times or values as a whole do not fit or cannot be relied on;
  flagged_unreliable additionally needs the distrusted part to be set aside.
- GoldStein values often fail to match an article's tone in every condition,
  FULL included; label what the model said and did, not whether it was right
  to distrust them.
- Judge the decisive method (for a run that did not answer, its final
  stretch), but a statement of distrust made earlier counts if the decisive
  method then avoids the values because of it.

---------------------------------------------------------------------------
3. terminal_state - for runs that did not answer; otherwise not_applicable
---------------------------------------------------------------------------

What the trace was doing when the 22,000-token ceiling cut it off.

still_working
  It was still generating new analysis - new assignments, new checks, new
  hypotheses - up to the cut, without having stated a conclusion it then
  abandoned.
answer_derived_then_reopened
  At some point it stated a letter as its conclusion (not as a candidate under
  test), then reopened the question to re-verify or reconsider, and was cut off
  while doing so.
answer_reached_output_truncated
  Its final stretch commits to a letter and it was cut off while writing up
  that conclusion or the final answer.
degenerated_into_repetition
  Its final stretch is dominated by repetition - removal markers, or the same
  check cycled with only cosmetic changes - with no substantive new steps.

If the final stretch is dominated by repetition, choose
degenerated_into_repetition even if a letter was stated earlier. For a run that
answered, give not_applicable.

---------------------------------------------------------------------------
4. Evidence and the rest
---------------------------------------------------------------------------

method_quote
  A verbatim quotation, at most 300 characters, from the step that decided the
  run (or from its final stretch, if it did not answer). Copy it exactly as it
  appears in the trace: same characters, same spacing, no ellipses inside, no
  paraphrase, nothing from a removal marker.
timestamp_quote
  A verbatim quotation, at most 200 characters, showing how the decisive method
  treated the temporal material. For absent, give an empty string.
terminal_quote
  For a run that did not answer, the last substantive sentence before the cut,
  verbatim, at most 200 characters. For a run that answered, an empty string.
last_named_option
  For a run that answered, its final letter. For a run that did not, the letter
  the trace most recently named as its current choice or leaning, or "none" if
  it never named one. Do not infer a letter it did not name.
leaning_method
  One of the eight decisive_method categories, or "none". It answers a
  different question from decisive_method: not what the run was doing when it
  stopped, but where its most recent leaning came from.
  - For a run that answered: the same value as decisive_method.
  - For a run that did not answer: the method of the step at which the trace
    named last_named_option as its choice or leaning - the line of reasoning
    that letter came out of, judged by the same rules as decisive_method. It
    may differ from decisive_method, for example when the run leaned towards a
    letter on recalled dates and then spent its final stretch rebuilding the
    CAMEO table. If last_named_option is "none", give "none".
leaning_quote
  For a run that did not answer and has a leaning, a verbatim quotation, at
  most 200 characters, from the step where that letter was named as the
  leaning. Otherwise an empty string.
confidence
  A number between 0 and 1: how sure you are of decisive_method.
\end{promptbox}

\paragraph{Condition-blind timestamp handling (G4).}
\texttt{\{\{RUNS\}\}} is replaced by the trace files of one batch.

\begin{promptbox}
You are labelling reasoning traces written by GLM-5.3-Flash on a multiple-choice
benchmark. You label one thing only: how the model treated the times listed in
its prompt. Read the background first.

BACKGROUND

Each run is one attempt at a question built from GDELT news events. The model is
given 6 to 8 event summaries, labelled (1), (2), (3) and so on, and asked to put
them in the order they actually happened, choosing between four candidate
orderings (A-D). It is also given a GoldSteinScale time series (one value per
timestamp) and a list of event times. The times are either calendar dates or
hours counted from the series' first record.

The prompt tells the model that the list of event times is "presented in a
random order" and does not say which time belongs to which event. Pairing times
with events is the model's job.

In some runs the listed times or the series may have been altered; you are not
told which runs, and you must not try to guess. Label only what the model wrote
and did. Whether its doubts were justified is not your question.

About the traces:
- Some runs hit the 22,000-token ceiling and end mid-thought; the header says
  which. For a run that answered, its final message follows a line reading
  FINAL MESSAGE.
- Pure repetition has been removed and replaced by a marker in square brackets.
  A marker is not the model's text.

THE TIMES LINK HAS TWO HALVES

The model can use the listed times in two ways:
- VALUES: it reads the GoldSteinScale value at a listed time and takes it as
  the value of the event that happened then (for example "time X has -10, that
  fits the attack").
- DATES: it takes a listed time as the moment one of the events happened (for
  example it places a date it recalls, or a date written in a summary, onto a
  listed time; or it uses a listed time to fix when an event occurred).

THE DECISIVE STEP

For a run that answered: the step its final letter came out of. For a run that
did not answer: what its final stretch was resting on. Judge the *_used fields
on this step only.

FIELDS

values_flagged (true/false)
  true if the model states that the series values at the listed times, as a
  whole, do not fit the events or cannot be relied on: they look scrambled,
  random, contradictory, noisy, "don't match", "not useful", "misleading",
  and the like. A remark that one or two values are surprising, followed by
  carrying on with the values, is false. A doubt about the model's own CAMEO
  lookup or memory is false.
values_used (true/false)
  true if the decisive step rests on values read at the listed times.
dates_flagged (true/false)
  true if the model states that the listed times do not fit when the events
  happened: a date it recalls or reads in a summary has no matching time in
  the list, the times seem wrong for these events, the times look scrambled,
  shifted or fabricated, and the like. That the list is in random order is
  stated by the prompt and is false. A single date slightly off a listed time,
  followed by placing the event there anyway, is false. A doubt about the
  model's own memory ("maybe I misremember") is false.
dates_used (true/false)
  true if the decisive step places events on listed times as their moments,
  or otherwise relies on a listed time being when an event happened.
values_quote
  If values_flagged, the verbatim sentence (at most 200 characters) where the
  model says so; otherwise "".
dates_quote
  If dates_flagged, the verbatim sentence (at most 200 characters) where the
  model says so; otherwise "".
used_quote
  A verbatim quotation (at most 200 characters) from the decisive step showing
  what it rested on.
confidence
  0 to 1, how sure you are of the four booleans together.

Every quote must be copied exactly from the trace: same characters, same
spacing, no ellipses, nothing from a removal marker.

{{RUNS}}
\end{promptbox}

\subsection{GDELT Category Definitions}
\label{app:gdelt-categories}

The figures name the eight categories of G3 as follows. List position and Guess are not shown in Figure~\ref{fig:accuracy} because no run uses them.

\paragraph{Date $\rightarrow$ slot} (\texttt{external\_calendar\_slot\_matching}).
The run recalls when a story happened and places it on one of the listed times.

\paragraph{Date $\rightarrow$ order} (\texttt{external\_direct\_chronology}).
The run orders the stories from remembered dates and ignores the listed times.

\paragraph{Value tone} (\texttt{semantic\_value\_matching}).
The run matches each story's tone, cooperative or conflictual, to the GoldStein values at the listed times.

\paragraph{CAMEO lookup} (\texttt{cameo\_lookup\_matching}).
The run tries to recall the exact GoldStein weight of each story's CAMEO action code.

\paragraph{Search series} (\texttt{series\_occurrence\_search}).
The run looks for likely values across the whole series and orders the stories by the positions it picks.

\paragraph{Date in summary} (\texttt{intrinsic\_text\_temporal\_constraints}).
A date or a time phrase written in a summary rules out options.

\paragraph{List position} (\texttt{positional\_alignment\_assumption}).
The run assumes that the $k$-th listed time belongs to event $(k)$.

\paragraph{Guess} (\texttt{unsupported\_fallback\_pick}).
The run picks a letter with nothing left to support it.

\paragraph{No leaning.}
The run hits the token limit without naming any option.

\subsection{Label Agreement}
\label{app:agreement}

The second labeller is Claude Sonnet~5. To check the labels, it labels a stratified sample again with the same prompt, without seeing the first labels.
Table~\ref{tab:agreement} gives its agreement with the first labels (Claude Opus~5 for NBA, Claude Opus~5.5 for GDELT).
For P3 the sample is 10 runs per condition, and both passes are condition-blind.
For G3 it is 6 runs per condition and per answered or truncated, in \textsc{Full}, \textsc{Relative} and \textsc{Shuffled}; G4 uses a new sample drawn the same way, and P4 uses the 40 runs of P3.
P3 has a lower $\kappa$ at 75\% agreement because one method, reading the score change, covers about 60\% of the runs.
For G4, the two labellers flag the same number of runs, with three disagreements in each direction, so the shares in Section~\ref{sec:experiments} do not depend on the labeller; the disagreements are borderline doubts about single values or dates.
For P4, both disagreements are runs that call a gap consistent with the chosen option without ruling another out.

\begin{table}[h]
  \centering
  \caption{Agreement between two labelling models on stratified samples. $\kappa$ is Cohen's kappa.}
  \label{tab:agreement}
  {\footnotesize\renewcommand{\arraystretch}{1.2}%
  \begin{tabular}{llccc}
    \toprule
    Task & Label & Runs & Agreement & $\kappa$ \\
    \midrule
    NBA   & final method (P3)         & 40 & 30/40 & 0.61 \\
    NBA   & first-pass method (P2)    & 40 & 36/40 & 0.86 \\
    NBA   & timestamp handling (P2)   & 40 & 38/40 & 0.91 \\
    GDELT & method in the figures (G3) & 36 & 29/36 & 0.75 \\
    GDELT & timestamp handling (G3)   & 36 & 32/36 & 0.59 \\
    GDELT & terminal state (G3)       & 36 & 34/36 & 0.92 \\
    NBA   & lag computed (P4)         & 40 & 40/40 & 1.00 \\
    NBA   & lag used (P4)             & 40 & 38/40 & 0.64 \\
    GDELT & times flagged (G4)        & 36 & 30/36 & 0.47 \\
    \bottomrule
  \end{tabular}}
\end{table}

\subsection{TCDP Audit}
\label{app:tcdp-audit}

Table~\ref{tab:tcdp-audit} applies TCDP to the two tasks.
For each rule of Figure~\ref{fig:tcdp}, it records whether the benchmark makes the rule necessary and, where it does, whether the trajectories follow it.
Of the ten rules, the two tasks exercise three (2c, 3a, 3d).

\begin{table*}[t]
  \centering
  \caption{\textbf{TCDP audit of the two tasks.}
  \emph{Not exercised}: the benchmark supplies a substitute (event order or the marked row), so the rule is never needed.
  \emph{Not measured}: our labels do not record it.
  Shares are \texttt{GLM-5.3-Flash} runs (Figures~\ref{fig:accuracy} and~\ref{fig:nba-methods}, Tables~\ref{tab:nba-lag} and~\ref{tab:agreement}).}
  \label{tab:tcdp-audit}
  {\footnotesize\renewcommand{\arraystretch}{1.25}%
  \begin{tabularx}{\textwidth}{l>{\raggedright\arraybackslash}X>{\raggedright\arraybackslash}X}
    \toprule
    Rule & NBA & GDELT \\
    \midrule
    1a Precedence from timestamps & Not exercised: possession chains move from the clock (14\%) to row order (13\%) at no cost & Not exercised: no question needs a causal claim \\
    1b Replay in timestamp order & Not measured & Not measured \\
    1c Lags from timestamps & Rarely used: no question needs a lag; 2\% use one under \textsc{Full}, 13\% under \textsc{Shuffled}, from moved times & Not exercised \\
    2a Target located by time & Not exercised: the table marks the row; 61--66\% read the score change there & Not applicable: no target moment \\
    2b Regimes from timestamps & Not measured & Not measured \\
    2c Cited events fit their times & Not measured & Held in \textsc{Full}: recalled dates placed on listed times (46\%); violated in \textsc{Shuffled}: still 42\%, on times that are not the events' \\
    3a Change tied to matching event & Not exercised: the table is pre-aligned & Violated in \textsc{Shuffled}: 35\% match tone to values at the wrong times \\
    3b, 3c Explanation and effect size & Not measured & Not measured \\
    3d Conflicts detected & Violated: 1\% notice under \textsc{Shuffled} & Violated: 36\% notice under \textsc{Shuffled} (11\% under \textsc{Full}) \\
    \bottomrule
  \end{tabularx}}
\end{table*}

\subsection{Representative Reasoning Trajectories}
\label{app:reasoning-trajectories}

Tables~\ref{tab:nba-trajectories} and~\ref{tab:gdelt-trajectories} give one representative \texttt{GLM-5.3-Flash} trace for each method.

\begin{table}[t]
\centering
\caption{\textbf{Representative NBA trajectories by method.} Each row gives excerpts from one GLM-5.3-Flash trace, in the order they appear; the excerpts are verbatim and [...] marks a cut. The step that decides the answer is marked $\blacktriangleright$; grey excerpts are context. \texttt{full/65} and \texttt{relative/65} are the same question. In \texttt{shuffle/37}, the event listed at 4:22 is a Team B basket that \textsc{Shuffled} moved there; in \textsc{Full} it is a Team A rebound. The run explains the mismatch away and answers wrong.}
\label{tab:nba-trajectories}
\small
\renewcommand{\arraystretch}{1.15}
\begin{tabularx}{\textwidth}{@{}>{\raggedright\arraybackslash}p{3.0cm}>{\raggedright\arraybackslash}X@{}}
\toprule
Method & Excerpts \\
\midrule
\meth{Read score change}{full/7}{Reads the two bracketing scores and stops.} &
\ctx{Past events: at 3:47-3rd, Player 13 (Team A) misses driving floating jump shot. At 3:45-3rd, Player 3 (Team B) defensive rebound.}
\dec{Target at 3:40-3rd: score goes from 75-75 to 75-78, so Team B scored 3 points.}
\ctx{Team A score remains 75, Team B goes from 75 to 78. So Team B made a 3-point shot.} \\
\midrule
\meth{Basketball knowledge}{full/95}{The score leaves two options; who is on the court decides.} &
\ctx{So Team B scored exactly 3 points between 9:52 and 9:47.}
\ctx{Both C and D are 3-pointers by Team B. Need to determine which player.}
\dec{So at 9:52-3rd, Team B on court: 5, 6, 9, 13, 15. Player 19 and Player 20 are NOT on the court.} \\
\midrule
\meth{Notices conflict}{shuffle/37}{Sees text and table disagree, explains it away, answers wrong.} &
\dec{Wait, the time series at 4:22 shows 111-108, and the event at 4:22 says Player 7 (Team B) makes 12-foot two point shot. Hmm, but the score updates might lag.}
\ctx{Actually at 4:14-4th the score is 111-110, which means Team B made the 2-point shot from 4:22 (that would make it 111-110).}
\ctx{The missing event at 4:14-4th should be a Team A score of 2 points (making it 113).} \\
\midrule
\meth{Clock + possession}{full/65}{Uses the next play's clock to see who kept the ball.} &
\ctx{So after the missed free throw, there should be a rebound.}
\dec{Then Team B had the ball again at 28.1 (Player 2 Team B misses 29-foot three). So Team B retained possession - meaning the rebound of the missed free throw went to Team B (offensive rebound).}
\ctx{So the missing event is likely an offensive rebound by Team B.} \\
\midrule
\meth{Order + possession}{relative/65}{The same question and chain, with offsets in place of clocks.} &
\ctx{At -1.7 seconds: Player 6 (Team B) makes 3-foot two point shot [...] Player 6 (Team B) misses free throw 1 of 1}
\dec{If it were option A (Player 4 lost ball turnover, steal by Player 9 A), possession would go to Team A, and the next event would be Team A offense, but at +1.4 it's Player 2 (Team B) missing a three, meaning Team B had the ball.}
\ctx{So the missing event must be a Team B offensive rebound: Player 4 (Team B) offensive rebound (option B).} \\
\bottomrule
\end{tabularx}
\end{table}

\begin{table}[t]
\centering
\caption{\textbf{Representative GDELT trajectories by method.} Same format as Table~\ref{tab:nba-trajectories}. \texttt{full/3} is the question of Figure~\ref{fig:gdelt-example}. The four answered examples are correct; across all runs, value-tone matching is right in only 41\% of \textsc{Relative} runs.}
\label{tab:gdelt-trajectories}
\small
\renewcommand{\arraystretch}{1.15}
\begin{tabularx}{\textwidth}{@{}>{\raggedright\arraybackslash}p{3.0cm}>{\raggedright\arraybackslash}X@{}}
\toprule
Method & Excerpts \\
\midrule
\meth{Date $\to$ slot}{full/3}{Recalls when each story ran and puts it on a listed time.} &
\ctx{(5) UK-Australia free trade deal agreement with 15-year cap on tariff-free imports. The broad agreement was announced June 15, 2021. [...] So (5) could be 2021-06-15 14:00:00.}
\ctx{(7) JBS cyberattack. JBS attack occurred May 30, 2021; Australian facilities affected, workers sent home around June 1, 2021. Timestamp 2021-06-02 05:45:00 fits (GoldSteinScale 3.0).}
\dec{Given (5) = June 15 (strong anchor) and (7) = June 2 (strong anchor) and (3) = July 9 (strong anchor), the answer must be D} \\
\midrule
\meth{Date $\to$ order}{no\_ts/142}{No times to use, so it orders the stories from memory.} &
\ctx{But we don't have explicit timestamps---only positions 1-117 with GoldSteinScale values.}
\ctx{I think the practical approach: use content knowledge of the news events' dates.}
\dec{Order: (3) [March/April] -> (4) [27 April] -> (5) [May] -> (7) [May] -> (2) [mid-June] -> (1) [late June] -> (6) [December]. = Option A.} \\
\midrule
\meth{Value tone}{relative/147}{Matches each story's tone to a GoldStein value.} &
\ctx{GoldSteinScale ranges from -10 (most conflictual) to +10 (most cooperative). So we can match each event's described tone to the Goldstein value at each timestamp.}
\ctx{Option A: (3)=0.0 [...] (7)=-6.5 (Prince William positive = -6.5?? No, mismatch). So A unlikely.}
\dec{Option D seems most consistent with sentiment matching: (6) protests = -6.5, (2) subsidy scheme = 7.4, (5) youth group = 7.0, (7) Prince William = 7.0} \\
\midrule
\meth{Date in summary}{full/141}{A date written inside one summary rules options out.} &
\ctx{(3) NFU Scotland hustings on January 22, 26, 28 - article likely published before Jan 22, so around Jan 21?}
\dec{But (3) mentions January 22, 26, 28 hustings - can't be Feb 9. So A wrong.}
\ctx{(3) at Mar 29 impossible (hustings in January). So B wrong. [...] (3) at Feb 9 impossible. So C wrong.} \\
\midrule
\meth{No leaning}{no\_ts/62}{Tries to rebuild the CAMEO score table and runs out of tokens.} &
\ctx{The idea is that each event matches a unique GoldSteinScale value in the series, and the position gives the chronological order.}
\ctx{Let me recall the CAMEO Goldstein table more carefully:}
\dec{Hmm, I'm fairly confident Accuse = -5.0? Hmm, hmm.} \\
\bottomrule
\end{tabularx}
\end{table}

\section{Full GDELT Results}
\label{app:gdelt}

Table~\ref{tab:gdelt-accuracy} gives accuracy on all 160 GDELT questions; Figure~\ref{fig:accuracy} gives the share of every method of \texttt{GLM-5.3-Flash}. The accuracies of individual methods cited in Section~\ref{sec:experiments} are computed over the runs that use them: Date$\to$slot is right in 62\% of its \textsc{Full} runs and 44\% of its \textsc{Shuffled} runs, and Value tone in 41\% of its \textsc{Relative} runs.
In a paired bootstrap over questions, \textsc{Relative} costs \texttt{GLM-5.3-Flash} 0.119 against \textsc{Full} (95\% interval $[0.025, 0.212]$), and \textsc{Shuffled} costs 0.200 ($[0.113, 0.287]$).
\texttt{Qwen3.6-35B-A3B} is near chance in every condition (0.24--0.29; always answering B or D scores 0.30), so the GDELT contrasts cannot be read for this model.

\begin{table}[h]
  \centering
  \caption{Accuracy on GDELT under the four timestamp conditions. Super- and subscripts give the distance to the upper and lower bounds of the 95\% bootstrap interval over questions. Responses without a valid answer count as wrong. Chance is 0.25.}
  \label{tab:gdelt-accuracy}
  {\footnotesize\setlength{\tabcolsep}{2pt}\renewcommand{\arraystretch}{1.25}%
    \begin{tabular}{lccc}
    \toprule
     & GLM-5.3-Flash & Sonnet-5 & Qwen3.6-35B-A3B \\
    \midrule
        FULL     & \cic{53}{0.519}{.081}{.081} & \cic{70}{0.656}{.075}{.075} & \cic{26}{0.294}{.069}{.069} \\
    RELATIVE & \cic{39}{0.400}{.075}{.075} & \cic{57}{0.550}{.075}{.075} & \cic{19}{0.238}{.069}{.063} \\
    NO-TS    & \cic{5}{0.125}{.056}{.050}  & \cic{42}{0.431}{.075}{.075} & \cic{25}{0.288}{.075}{.069} \\
    SHUFFLED & \cic{29}{0.319}{.075}{.069} & \cic{62}{0.594}{.075}{.075} & \cic{20}{0.250}{.069}{.069} \\
    \bottomrule
  \end{tabular}}
\end{table}

\section{Limitations and Future Work}
\label{app:limitations}

Our study has several limitations. First, we evaluate only two TSRBench tasks (NBA: 150 questions, four LLMs; GDELT: 160 questions, three LLMs). The reliance on sequence order may differ across domains, temporal granularities, model families, and benchmark designs; broader evaluations (e.g., finance, healthcare, weather) are needed.

Second, TCDP is only partially exercised: neither task requires lag reasoning, and only 2\% of NBA runs under \textsc{Full} use a lag (Appendix~\ref{app:tcdp-audit}). More generally, TCDP checks temporal consistency but does not validate causal correctness; we make no claim of causal identification.

Third, trajectory analyses cover only GLM-5.3-Flash. Labels are produced by LLM labellers (Appendix~\ref{app:reasoning-procedures}) with moderate agreement (NBA method: $\kappa = 0.61$; GDELT misalignment: $\kappa = 0.47$). Some labels cannot be exactly regenerated (Claude Opus~5 is unavailable), and verbalized reasoning may not reflect internal processes; thus our shortcut claims rest mainly on the accuracy interventions.

Fourth, GDELT performance partly reflects memorized real-world dates despite the prompt restriction, and may change for events after a model's training cutoff.

Finally, our interventions operate in settings where event--series alignment is largely given by the benchmark. Stronger benchmarks should require models to recover alignment from independently retrieved or asynchronously observed evidence.

\paragraph{Broader impact.}
This is an evaluation study. Its main expected benefit is to reduce over-trust in LLM-based analysis of time series in sensitive settings such as clinical or financial monitoring, by showing that high benchmark accuracy need not reflect temporal grounding. We do not identify direct negative impacts beyond those generally associated with LLMs.


\end{document}